\documentclass[sigconf]{acmart}

\usepackage{xcolor}
\usepackage{caption}
\usepackage{subcaption}
\usepackage{algorithm}
\usepackage{algorithmicx}
\usepackage{algpseudocode}
\usepackage{wrapfig}
\usepackage{etoolbox}
\usepackage{hyperref}
\usepackage[inline]{enumitem}

\usepackage{tikz,pgfplots}
\usetikzlibrary{plotmarks,shapes,shapes.arrows,positioning, backgrounds}

\newrobustcmd*{\mytriangle}[1]{\tikz{\filldraw[draw=#1,fill=#1] (0,0) --
(0.2cm,0) -- (0.1cm,0.2cm);}}

\usepackage{graphicx} 
\usepackage{subcaption}
\usepackage{mathtools}
\usepackage{amsmath,amsfonts}
\usepackage{dirtytalk}
\usepackage{bm}
\usepackage{float}
\newbox{\bigpicturebox}
\definecolor{blue_online}{RGB}{4,120,177}

\algnewcommand{\LineComment}[1]{\State \(\triangleright\) #1}
\AtBeginDocument{%
  \providecommand\BibTeX{{%
    \normalfont B\kern-0.5em{\scshape i\kern-0.25em b}\kern-0.8em\TeX}}}

\copyrightyear{2022} 
\acmYear{2022} 
\setcopyright{acmlicensed}\acmConference[KDD '22]{Proceedings of the 28th ACM SIGKDD Conference on Knowledge Discovery and Data Mining}{August 14--18, 2022}{Washington, DC, USA}
\acmBooktitle{Proceedings of the 28th ACM SIGKDD Conference on Knowledge Discovery and Data Mining (KDD '22), August 14--18, 2022, Washington, DC, USA}
\acmPrice{15.00}
\acmDOI{10.1145/3534678.3539196}
\acmISBN{978-1-4503-9385-0/22/08}

\begin{document}

\title{Human-in-the-Loop Large-Scale Predictive Maintenance of Workstations}

\author{Alexander Nikitin}
\email{alexander.nikitin@aalto.fi}
\affiliation{%
  \institution{Department of Computer Science, Aalto University}
  \city{Espoo}
  \country{Finland}
}

\author{Samuel Kaski}
\email{samuel.kaski@aalto.fi}
\affiliation{%
  \institution{Department of Computer Science, Aalto University}
  \city{Espoo}
  \country{Finland}
}
\affiliation{%
  \institution{Department of Computer Science, University of Manchester}
  \city{Manchester}
  \country{UK}
}

\renewcommand{\shortauthors}{Alexander Nikitin \& Samuel Kaski}

\begin{abstract}
    Predictive maintenance (PdM) is the task of scheduling maintenance operations based on a statistical analysis of the system’s condition. We propose a human-in-the-loop PdM approach in which a machine learning system predicts future problems in sets of workstations (computers, laptops, and servers). Our system interacts with domain experts to improve predictions and elicit their knowledge. In our approach, domain experts are included in the loop not only as providers of correct labels, as in traditional active learning, but as a source of explicit decision rule feedback. The system is automated and designed to be easily extended to novel domains, such as maintaining workstations of several organizations. In addition, we develop a simulator for reproducible experiments in a controlled environment and deploy the system in a large-scale case of real-life workstations PdM with thousands of workstations for dozens of companies.
\end{abstract}

\ccsdesc[500]{Computing methodologies~Machine learning}
\ccsdesc[500]{Applied computing}
\ccsdesc[300]{Information systems~Data mining}

\keywords{machine learning, human-in-the-loop, predictive maintenance, \\ 
\mbox{Bayesian optimization}, applications}

\maketitle

\section{Introduction}
Predictive maintenance (PdM) determines the optimal timepoint for maintenance actions based on the condition of equipment. This task is becoming one of the most critical problems in systems management. For example, within the Industry 4.0 concept \cite{chung2016internet}, PdM is described as an essential part of computer-assisted manufacturing. PdM techniques have been applied in several domains, including manufacturing \cite{pdm_for_manufacturing}, the Internet of Things \cite{civerchia2017industrial}, and air force systems \cite{gravette2015achieved}.

The development of effective PdM systems for workstations is driven by the growth in the number of personal computers, laptops, and servers. In this application, PdM systems will enhance the efficiency of maintenance engineers and improve customer experience. However, while predictive maintenance approaches have been used in many fields (see Sec.~\ref{section:background} for a more detailed discussion), to the best of our knowledge, no studies have proposed and field-tested machine learning (ML) solutions to the PdM of workstations.

We study the predictive maintenance task through the lens of human-in-the-loop (HITL) machine learning methods. HITL ML is a set of ML techniques that interact with human experts in the decision-making loop. HITL methods produce various benefits, including better predictive performance \cite{monarch2021human} and domain adaptation \cite{nikitin2021decision}. These approaches are especially beneficial for safety-critical tasks, such as airborne collision avoidance \cite{li2014synthesis} and surgery automation \cite{fosch2021human}. The success of prior works in applying HITL approaches to critical tasks and the ability of HITL to assist in decision-making indicate the potential for employing these methods in PdM.

\textbf{Contributions.} In this paper, we: \emph{(i)} formulate the predictive maintenance problem for workstations, \emph{(ii)} propose ML methods for workstation PdM, \emph{(iii)} explain how domain experts can be effectively included in the PdM process using HITL machine learning approaches, and \emph{(iv)} describe a field-tested large-scale implementation of our approach and motivate the system design choices.

\textbf{Reproducibility.} We validate the developed system using both offline and online. For offline validation, we measure the model's performance on historical data and develop a simulator. The simulator allows for generating datasets that mimic properties of the real data and allows for testing the methods in a controllable environment where the generation process parameters can be varied. The source code of the toy experiments with synthetic data and the simulator are available at \url{https://github.com/AaltoPML/human-in-the-loop-predictive-maintenance}. Also, we describe the experiments with historical data collected from one thousand workstations to validate the models and online validation of the system in the field.

\textbf{Application benefits.} This work is important for several parties: maintenance engineers/domain experts, companies with extensive IT infrastructure, and machine learning researchers. For domain experts, the work proposes a tool that allows them not only to prevent upcoming malfunction and automatize routine parts of their work but also provides a user-friendly, interactive, and reliable assistant. For businesses, this tool allows cost-efficient workstation monitoring. It consequently leads to fixing possible malfunctions in advance, without hurting the user experience. Lastly, for machine learning researchers, this application shows an example of effective interaction with humans in PdM applications and, importantly, a reproducible way to simulate and test their methods on synthetic PdM data.

\section{Background}
\label{section:background}
\emph{Predictive maintenance.} Machine learning approaches to PdM include a variety of methods ranging from knowledge-based techniques to deep learning (DL) systems \cite{Ran2019ASO}. Recent developments have primarily focused on DL approaches, including autoencoders \cite{ren2018bearing}, convolutional neural networks \cite{yang2019remaining}, and recurrent neural networks \cite{guo2017recurrent}. Knowledge-based systems were popular in the early days of machine learning, but, currently, they have given way to data-driven approaches. In terms of applications, predictive maintenance methods have often been applied to high-cost systems, including ships, aircraft fleets, and turbofans \cite{Ran2019ASO}, \cite{saxena2008turbofan}. However, with the increase of data available for more standard systems, the scope is widening to encompass more common problems, such as those related to the Internet of Things \cite{distributed_iot_pdm}, and ATMs \cite{ATM_event_logs}. Several approaches to the analysis of the tickets in IT services were developed, for example, \cite{zhou2017star} and \cite{potharaju2013juggling}. These approaches focus on extracting useful information from the textual description and otherwise improve reaction time; they do not tackle the proactive incident management aspect, which is the main focus of our work. Scalability and deployment of the methods in changing environments remain open challenges for all application areas in predictive maintenance, and we resolve parts of them in this work.

\emph{Human-in-the-loop machine learning.} Machine learning agents can gain useful information from interactions with humans (HITL methods). These methods have resulted in many advantages, including performance improvements \cite{amershi2014power}, better explainability \cite{celino2020explanation}, and tackling computationally hard problems \cite{holzinger2016interactive}. Humans may assist the learning process in various ways, for example, by including inductive biases in building kernel machines \cite{wilson2015human}. In this work, we show how ML models can interact with experts in the PdM process via explicit decision rule feedback.

\emph{Recommender systems.} Recommender systems are machine learning methods used to predict the relevancy of a given collection of items to a group of users. Recommender systems have been applied in many other domains, including music \cite{yoshii2008efficient}, and research articles \cite{beel2013research}. Nevertheless, they have been applied to PdM only seldom. For example, in \cite{das2013maintenance}, the authors showed how to recommend maintenance actions for an industrial piece of equipment. We utilize a recommender engine with additional information to represent the models' results.

\section{Predictive maintenance of workstations}
We consider the task of PdM of a set of workstations: personal laptops, servers, and other computers. Our goal is to develop a system that will predict whether a problematic situation will occur in the future and guide domain experts to prevent it. An essential practical requirement is that the system should be applicable to different organizations' workstations (in ML terms, automatically adapt to novel domains). Specifically, we need to deal with dozens of organizations with thousands of workstations each. The development of the machine learning system consists of the following steps, described in more detail in Sec.~\ref{section:deployment}: extracting numerical measurements from the workstations, processing the extracted features and preparing an ML dataset, application of ML algorithm to the extracted dataset, deployment and continuous integration of the model in the PdM pipeline, and collecting feedback from domain experts. Importantly, we allow domain experts to give explicit decision rule feedback, which improves the model's predictive performance and allows experts to influence the system's behavior.

\subsection{Dataset}
\label{section:producing_ml_dataset}

\begin{figure}
\includegraphics[width=\columnwidth]{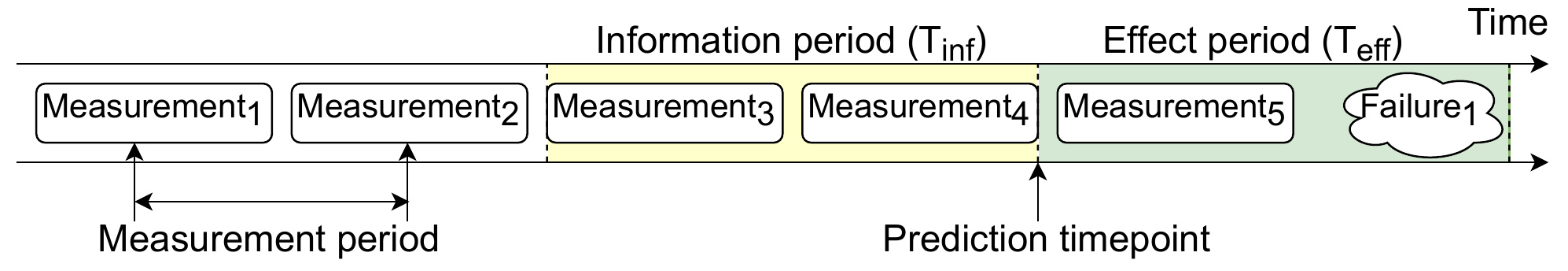}
\caption{The system analyzes aggregated measurements and predicts future (relative to prediction timepoint) failures. The information period is a time segment during which the data is aggregated. The effect period is a period in the future relative to the prediction timepoint, which is used to calculate the target variable (an indicator of a failure).} 
\label{figure:dataset_to_ML}
\end{figure}

Data collection can be performed using agents installed on each maintained computer. The agents submit their measurements once every measurement period (here two hours), resulting in a heterogeneous set of temporal measurements. For learning to predict failures in the future, we define two time intervals: \emph{information period} ($T_{\text{inf}}$) and \emph{effect period} ($T_{\text{eff}}$). Let us consider an arbitrary time point and an arbitrary workstation. We analyze a period before the prediction point (information period) to gather workstation measurements. Then, we make predictions of breakdowns in a specific time interval in the future (effect period). Schematically, the information and effect periods are shown in Fig.~\ref{figure:dataset_to_ML}. To construct an ML task out of these data, we need to specify a method for generating covariates (inputs to the predictor) and target variables.

\emph{Covariates.} We use several informative values that describe the condition of the workstations; these measurements include CPU load, used memory, installed drivers, and software monitoring alarms. The features of a particular workstation can be described as a  sequence of data points $\bm{x} \in \mathbb{R}^d$ for each timestamp, where $d$ is the number of numerical measurements that characterize the workstations. The dataset $\bm{X}$ is a collection of situations where each element $\bm{x}_{i} \in \bm{X}$ is a set of collected measurements over $T_{\text{inf}}$ periods. The data points are either aggregated over each period to multi-dimensional vectors $\bm{x}_i \in \mathbb{R}^{d}$ as described in Eq.~\ref{equation:feature_aggregation} or to time-indexed measurements $\bm{x}_{i} \in \mathbb{R} \times \mathbb{R}^{d}$. We aggregate the measurements over the information periods and extract statistics that we use as features, using the following approach:
\begin{equation}
    \label{equation:feature_aggregation}
    \bm{\Tilde{x}}_i = \underset{\text{agg} \in \mathcal{A}}{\oplus} \text{agg}\left(\bm{x}^{t-T_{\text{inf}}}_i, \ldots, \bm{x}^{t}_i\right),
\end{equation}
where $\mathcal{A}$ is a set of the aggregational operators, and $\oplus$ concatenates results into one vector. The aggregational operators can include, for example, percentiles, rolling mean, or mode. Later in the text, we will denote aggregated features as $\bm{x}_i$ and $\bm{X}$.

Some of the extracted information can be represented as sets; for example, the sets of installed drivers or software programs. We have applied a natural language processing (NLP) technique called tf-idf \cite{jones1972statistical} to transform temporal information about sets into covariates. Consider installed drivers as an example. We transformed the collected data on drivers to a list of pairs (name and version) and computed the standard measures term frequency (tf) and inverse document frequency (idf) as follows:
\begin{equation}
\begin{split}
    \text{tf}(d, \bm{x})& = \frac{\#\{d^{\prime} = d | d^{\prime} \in \bm{x}\}}{\#\{d^{\prime} = d | d^{\prime} \in \bm{x}^{\prime}, \bm{x}^{\prime} \in \bm{X}\}},\\ \text{idf}(d)& = \log{\frac{N}{\#\{ d \in \bm{x} | \bm{x} \in \bm{X}\}}}, \\ \text{tf-idf}(d, \bm{x})& = \text{tf}(d, \bm{x}) \times \text{idf}(d),
\end{split}
\end{equation}
where $d$ is a driver, $\bm{x}$ is an element of the training set, and $D$ is a whole set of drivers. With symbol $\#$, we denote the cardinality of a set, and $N$ is the size of the training dataset. We use the notation $d \in \bm{x}$ to denote that the driver $d$ was installed at some point in the $x$'s information period. The tf-idf representation of installed drivers is then concatenated to the vector with the rest of the features.

\emph{Targets.} The machine learning model $\mathrm{M}$ tries to predict whether a problem will occur in the following $T_{\text{eff}}$ relative to the prediction point. To indicate the health of a particular workstation, we count system errors and warnings (in this paper, we will call them alerts) for each workstation during the effect period. For example, some alerts indicate a problem loading the system (errors for slow startup) or report software's frequent crashes. To distinguish failures from normal functioning, we compare the total number of alerts with a percentage threshold, which results in a binary classification problem. From discussions with domain experts, we noticed that they considered the total number of errors and alerts to be a more fine-grained indication of problems than other approaches (e.g., counting unique types of alerts or user requests).

\subsection{Bayesian optimization of $T_{\text{inf}}$ and $T_{\text{eff}}$}
\label{section:bayesian_optimization}
In order to transform raw data into a machine learning dataset, it is crucial to select optimal information and effect periods. Let us consider function $\text{perf}(T_{\text{inf}}, T_{\text{eff}}, \mathrm{M}, D)$, which approximates the predictive performance of model $\mathrm{M}$ on dataset $D$ with $T_{\text{inf}}$ and $T_{\text{eff}}$ parameters. To find the optimal periods, we might take the brute-force grid search approach and evaluate $\text{perf}$ for a large set of values. That would, however, be very inefficient as we need to employ the system for different organizations and repeat this process. Moreover, to get statistically significant estimates of $\text{perf}$, we have to run computationally inefficient cross-validation for each combination of the parameters. Rather than adapting the brute-force approach, we optimize the function $\text{perf}$ with Bayesian optimization (BO) \cite{movckus1975bayesian}.

To employ the BO methods, it is necessary to define a surrogate function that will approximate the model's generalization performance. For this purpose, we define a Gaussian process (GP) prior over functions: 
\begin{equation}
f \sim \mathcal{GP}(\mu(\mathbf{x}), K(\mathbf{x}, \mathbf{x})).    
\end{equation}
GPs are described with a mean function $\mu(\bm{x})$ and a covariance function (kernel) $K$, such as squared exponential, periodic, and radial basis function (RBF) kernels \cite{duvenaud2014automatic}. In our work, we used the RBF kernel:
\begin{equation}
    K(\mathbf{x}, \mathbf{x}^{\prime})= \exp (-\gamma\left\|\mathbf{x}-\mathbf{x}^{\prime}\right\|^{2}),
\end{equation}
where $\gamma$ is a length scale hyperparameter.
For each iteration, the BO algorithm selects the next most informative point and evaluates the function at this point. This selection process is organised by the function called acquisition function. In our work, we used the upper confidence bound (UCB) acquisition function \cite{brochu2010tutorial} which allows the algorithm to balance between exploration and exploitation steps:
\begin{equation}
    a_{UCB} = m(\mathbf{x}) + \kappa \sigma(\mathbf{x}),
\end{equation}
where $\kappa$ is the exploration-exploitation tradeoff parameter and $\sigma(\bm{x}) = \sqrt{K(\bm{x}, \bm{x})}$. After choosing the point, the BO algorithm updates the GP posterior with the acquired value.

\section{Human-in-the-loop predictive maintenance}
\subsection{Base ML model}
\label{section:ml_for_workstations_pdm}
We employ a machine learning algorithm to predict whether a problem will occur in the following $T_{\text{eff}}$ time. More formally, our aim is to develop a binary classifier $\textbf{M}: \bm{X} \rightarrow \{0, 1\}$, where each element of $\bm{X}$ is an aggregated set of workstation measurements during $T_{\text{inf}}$ (yellow segment in Fig.~\ref{figure:dataset_to_ML}), and $0$ indicates normal functioning of a system and $1$ malfunctioning.

As a base ML model, we consider several approaches that have shown excellent results in ML problems with extensively heterogeneous data: gradient boosting \cite{friedman2001greedy}, and extremely randomized trees \cite{geurts2006extremely}. We also compared the results to logistic regression. These methods possess several advantages over deep learning approaches, such as long short-term memory (LSTM) \cite{hochreiter1997long}, and gated recurrent unit (GRU) \cite{cho-etal-2014-properties} neural networks. These advantages include the ability to cope with a large number of noisy and missing measurements and interpretability, the latter being a requirement for the machine learning method in our solution and being also important for other critical systems.

\subsection{Decision rule elicitation (DRE)}
\label{section:dre}
In this work, we propose to include the domain experts in the predictive maintenance loop in a novel way for PdM --- via explicit decision rule elicitation \cite{nikitin2021decision}. Decision rule elicitation provides several advantages over standard techniques: better domain adaptation, predictive performance, and explainability.

An informal survey among maintenance engineers showed that they were able to explicitly explain their decision-making process, justifying what heuristics they used to check whether a workstation was broken. Such explicit heuristics can be represented as decision trees or logical formulas, in which the experts often rely on the current warnings and alarms and compare measurements (e.g., hard disk free space) with some threshold values they have identified based on their earlier experience. An example of a decision rule is
\begin{align}
\left[
\begin{array}{ll}
    1, \text{if } c((a_{\text{perf}}, a_{\text{mem}}), \bm{x}) > 0 \vee c(a_\text{\text{soft}}, \bm{x}) > 1,\\
    0.5, \text{otherwise,}
\end{array}
\right .
\end{align}
where $a_{*}$ are alerts related to different aspects of the workstation's operation, in this example, performance, memory, and software, $x$ is an observation, and $c$ is a function that returns the number of alerts during the information period of the observation. The value one in the first equation means that the device has to be examined immediately, and 0.5 in the second means that there is not enough information to determine the workstation condition. These values can be interpreted as a subjective probability of malfunctioning.

Following the approach proposed in \cite{nikitin2021decision}, we query experts for decision rules when they discover a particular prediction to be wrong. The decision rule feedback is taken into account with the following model: Consider the explicit decision feedback rules from the user as a set of Boolean predicates, $f_1, f_2, \dots, f_n$ elicited from domain experts. We generalize this approach to return a discrete probability distribution over the target variable domain: $f: \bm{X} \longrightarrow \mathbf{P}$. In our case, $\mathbf{P}$ refers to the probability of a malfunction. The feedback rules were provided by domain experts in natural language and then manually translated into the set of functions. The resulting classifier can be written as:
\begin{equation}
\label{equation:feedback_classifier}
    C_{\text {feedback }}(\bm{x})=\zeta(\sum_{i=1}^{F} f_{i}(\bm{x}) \operatorname{sim}(\bm{X}_{\text {test }}^{i}, \bm{x})\,  \theta_{i}).
\end{equation}
Here $F$ is the total number of feedback rules, $\theta$ are weights of the rules that could be either learned from the data or manually selected using prior knowledge, $\operatorname{sim}$ is a similarity measure that is calculated between the datapoint where the feedback was given $\bm{X}_{\text {test }}^{i}$ and  $\bm{x}$ (the datapoint where the model is applied) and $\zeta$ is a smoothing function, for example linear (i.e., for averaging $\zeta(x) = x/F$) or sigmoid. Finally, we combine the feedback-based model with an ML model learned from data as a weighted average with weight $\alpha$:
\begin{equation}
\label{equation:decision_rule_elicitation_weighted_average}
C(\bm{x})=\alpha \mathrm{M}(\bm{x}, \bm{\theta}_{M})+(1-\alpha) C_{\text{feedback}}(\bm{x}, \bm{\theta}_{f}),
\end{equation}
where $\bm{\theta}_{M}$ is a vector of the parameters of the machine learning algorithm $M$, and $\bm{\theta}_{f}$ is a vector of parameters of the feedback model. $C(\bm{x})$ serves as the estimate of how problematic a particular situation is. This value is not a calibrated probability, but is useful for ranking the workstations.

\subsection{Multiple domains and decision rules}
In practice, it is necessary to answer two questions: \emph{(i)} when a new decision rule appears for a particular domain, should it be used in other domains? \emph{(ii)} when a new domain appears (a new set of workstations joined the system), which of the existing decision rules should we use? The first question can be addressed by comparing historical performance with and without this rule for each domain. This operation can be performed efficiently because the model and decision rule outputs on the historical data can be cached and reused; the rest can be parallelized.

A naive algorithm cannot effectively resolve the second problem. It would need to compare all the subsets of the collected decision rules, which will result in $O(2^{|R|} E)$ time complexity, where $|R|$ is the number of unique decision rules, and $E$ is the complexity of evaluating the decision rules on the training dataset and, even if an efficient algorithm existed, choosing from many alternatives without overfitting would require a big data set which may not always be available. Instead, we propose an approach that exploits similarities between domains using maximum mean discrepancy (MMD),
\begin{equation}
    \operatorname{MMD}(P, Q)=\left\|\mathbb{E}_{X \sim P}[\varphi(X)]-\mathbb{E}_{Y \sim Q}[\varphi(Y)]\right\|_{\mathcal{H}},
\end{equation}
where the distributions $P$ and $Q$ are defined over domain $\mathcal{X}$, and $\varphi$ is a feature map from $\mathcal{X}$ to Hilbert space $\mathcal{H}$ \cite{gretton2012kernel}. 

\textbf{Algorithm.} First, we sort all domains by the MMD similarity to the novel domain. Then, we iterate through the domains and add decision rules by batches for each domain if they increase predictive performance. Otherwise, we add the domains to queue $\rho$. Next, we iterate through the queue and add each decision rule individually if it increases the model's performance. The whole procedure is shown in Algorithm~\ref{alg:decision_rule_selection}.
\begin{algorithm}[t!]
   \caption{Decision rule selection for a novel domain.}
   \label{alg:decision_rule_selection}
\begin{algorithmic}
   \State {\bfseries Input:} new domain $d_\text{new}$, set of domains: $\mathcal{D}$, decision rules for each domain $R$, DRE model $M$, historical data for each domain $H$;\\
   The algorithm uses two external functions:
   \State {\bfseries Perf:} a function that takes a model, historical data, and the list of predicates. Returns: predictive performance (performs k-fold cross-validation).
   \State {\bfseries MMD:} takes two data samples and returns a measure of similarity between the distributions.\\
   
   \LineComment{Sort domains by MMD with $d_{\text{new}}$}
   \State $\mathcal{D}$ $\gets$ sort($\mathcal{D}$, key=$\lambda d [\text{MMD}(H[d_{\text{new}}], H[d])]$)\\
   
   \LineComment{Iterate over domains and add rules as a batch}
   \State $\rho \gets \emptyset$; \Comment{queue of not added domains for the next phase}; 
   \For{$d \in \mathcal{D}$}
        \State $p_\text{new} \gets \text{perf}(M, H[d_\text{new}], R[d_\text{new}])$;
        \State $p_\text{updated} \gets \text{perf}(M, H[d_\text{new}], R[d_\text{new}] \cup R[d])$
        \State $\delta \gets |p_\text{new} - p_\text{updated}|$;
        \If{$\delta > 0$}
            \State $R[d_\text{new}] \gets R[d_\text{new}] \cup R[d]$;
        \Else
            \State $\rho \gets \rho \cup d$;
        \EndIf
   \EndFor
   \\
   \LineComment{Add the rest of the rules in a greedy manner}
   \For{$d \in \rho$}
        \For{$r \in R[d]$}
        \State $\delta \gets \ldots$ \Comment{compare performance with and without $r$}; 
            \If{$\delta > 0$}
                \State $R[d_\text{new}] \gets R[d_\text{new}] \cup \{r\}$;
            \EndIf;
        \EndFor
   \EndFor
   
   \State {\bfseries Return:} $R$;
\end{algorithmic}
\end{algorithm}

\textbf{Complexity.} The proposed algorithm works with time complexity $O(E \max(|R|, |\mathcal{D}|))$.

\textbf{Assumptions and optimality.} For a fixed model $M$, we will call a decision rule $r$ \textit{aligned} with a domain $d_i$:  $r \sim d_i$, if exclusion of $r$ makes the whole model less accurate on the domain $d_i$. Algorithm~\ref{alg:decision_rule_selection} returns a quasi-optimal allocation of decision rules under the following assumptions: for a new domain $d_{\text{new}}$ and for processed domains $d_1, \ldots, d_k$, such that
\begin{multline}
   \text{MMD}(d_1, d_{\text{new}}) \leq \ldots \leq \text{MMD}(d_i, d_{\text{new}}) \leq \\ \ldots \leq \text{MMD}(d_k, d_{\text{new}}),
\end{multline}
there exists $j$, such that domains $d_1, \ldots, d_j$ are close to $d_{\text{new}}$ ($\forall r \in R[d_{1 \leq i \leq j}]: p(r \sim d_{\text{new}}) = 1$), and  $\forall r \in R[d_{j < i \leq k}]: p(r \sim d_{\text{new}}) = \epsilon$ for a small $\epsilon$ (the probabilities are calculated over all model specifications, including different combinations of decision rules from $R$). We can see that by the design of the algorithm we first add the rules, only if they are aligned with the novel domain, starting from the closest domains. By the construction of this algorithm, the model's performance does not decrease while adding decision rules from other domains. When the first loop is over, variable $\rho$ stores only the domains where all the rules were not aligned with $d_{\text{new}}$. By the second loop we add those that became aligned in combination with other domains. When we add the rest of the decision rules linearly, we neglect the higher-order of epsilon probabilities of combining several rules to be beneficial for generalization performance. Even though the assumptions are quite strong, we found them to be realistic and helpful in cold starting novel domains.

\subsection{Recommender engine for predictive maintenance}
\label{section:recommender_engine}

\begin{figure}
    \includegraphics[width=\columnwidth]{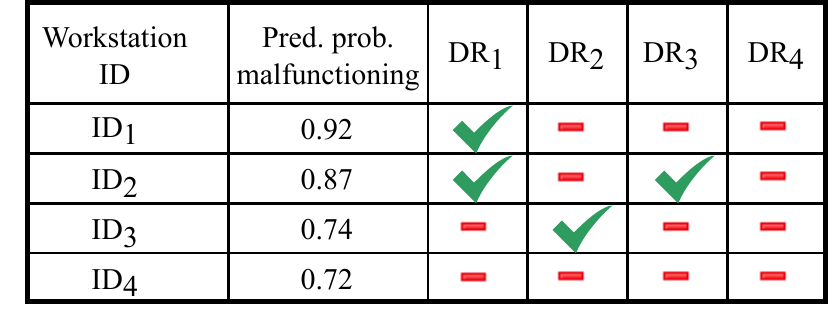}
    \caption{Schematic visualization of the outputs provided to the domain experts. The first column identifies the workstations, ordered by the model's predictions (the second column). The second column shows the (uncalibrated) probability of malfunctioning. Columns $DR_{i}$ are the results of elicited rules, selected by domain experts as the most interpretable.} 
    \label{figure:recommendation_example}
\end{figure}

We combine the results produced by the ML algorithm and elicited decision rules into a ranked list for domain experts. The system predicts the probability of a device malfunctioning (the second column in Fig.~\ref{figure:recommendation_example}) and shows the ranked list of the devices ordered by the probabilities. The outputs are enriched with additional information that helps experts to make faster decisions and navigate the list effectively. We used outputs of the decision rules as the additional information (columns named $DR_i$ in Fig.~\ref{figure:recommendation_example}). Domain experts can vary their load by selecting the number of predictions shown. Moreover, they can aggregate similar problems with SQL-like queries to navigate the list faster. An example of such query, to select only workstations with the probability of malfunction higher than 0.9 and with active decision rules 17 and 23, is 
\begin{verbatim}
    SELECT * FROM CUR_TICKETS WHERE 
                    P_M > 0.9 AND DR_17 AND DR_23 .
\end{verbatim}

\section{Industrial implementation}
\label{section:deployment}
\begin{figure}
    \includegraphics[width=\columnwidth]{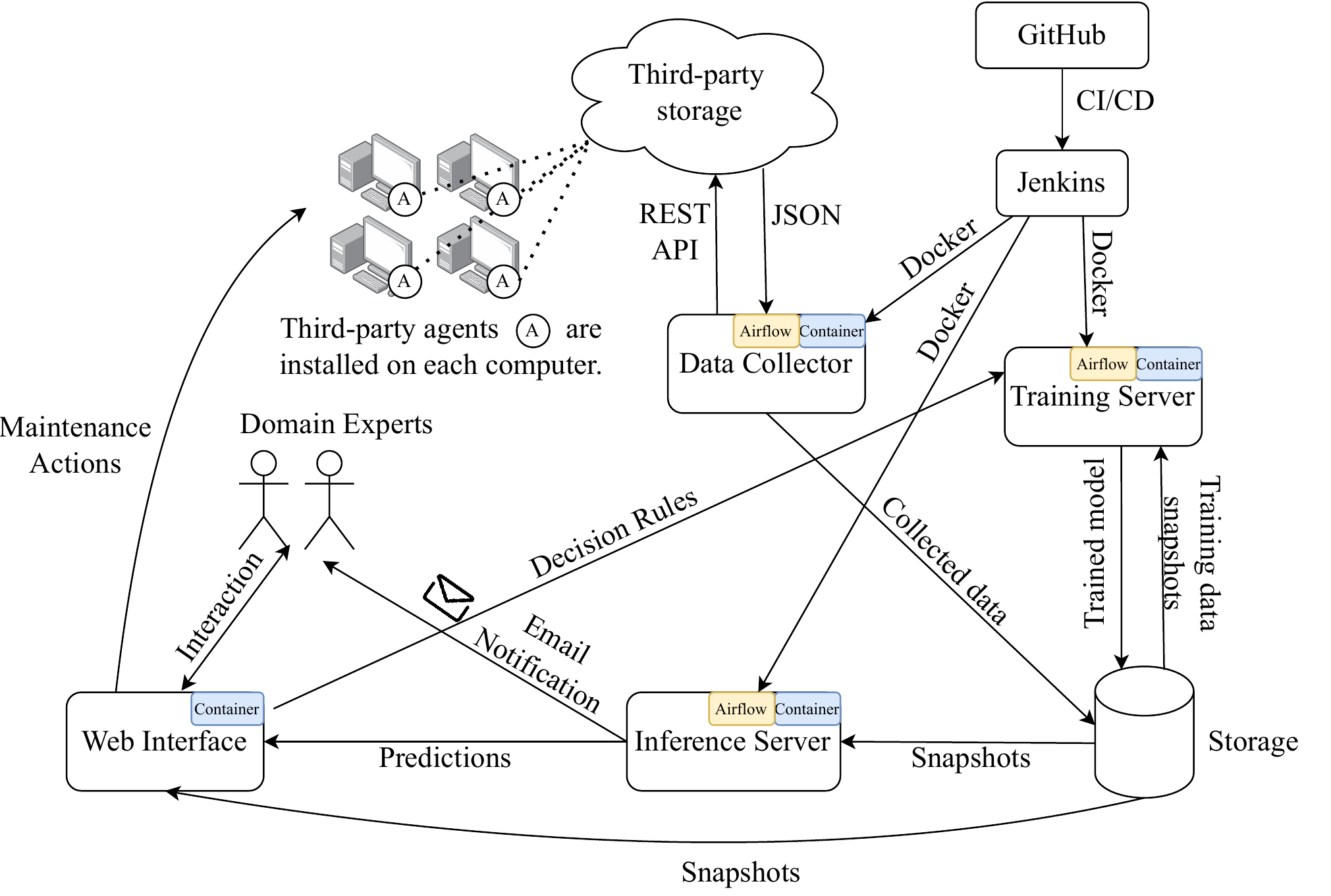}
    \caption{Interaction of the services and agents in the developed system. The system's end goal is to predict future workstation (on the left of the picture) problems and resolve them proactively. The system functioning starts from the source code storage (illustrated in the right upper corner). The code is built by Jenkins and distributed as Docker containers to the data collector, training, and inference services (in the middle of the image). The inference server notifies the domain experts via email notifications daily, and domain experts perform actions based on these predictions. The domain experts return their decisions to the training server and perform maintenance through the web interface.} 
    \label{figure:deployment_diagram}
\end{figure}
In practice, it is not enough to develop a model with decent predictive performance; it is necessary to build the whole system with continuous data delivery, fault tolerance, and effortless deployment of new versions to the end-users. Here, we describe the design of our system. The system consists of three principal components: a data collector, training, and inference services, as shown in Fig.~\ref{figure:deployment_diagram}. The process can be described as follows: Third-party software collects measurements from the workstations and stores them in the third-party cloud. Our data collector extracts the data from the REST API and puts it to S3-compatible storage that is implemented via Ceph\footnote{\url{https://ceph.io/}}. The data is stored in a raw format, roughly 1.2 Tb during a year. The second component (the training service) extracts and preprocesses data from S3, trains a machine learning model, and saves the serialized model to S3 storage. The inference service delivers predictions by running the machine learning model on the most recent data and returning the predictions to domain experts. Our system also provides the predictions to domain experts daily, and the training service runs as an Airflow job. The services are deployed as docker containers\footnote{\url{https://www.docker.com/}} that are orchestrated via Kubernetes \footnote{\url{https://kubernetes.io/}}. Docker containers are built by Jenkins server \footnote{\url{https://www.jenkins.io/}} where the new versions of code are delivered by GitHub continuous delivery. The model versioning is done via versioning the inference service codebase and changing the newly trained model's path. The same approach can be extended to creating a real-time service by horizontal scaling. Also, the analysis of the workstations is delivered via a web interface, where domain experts can give decision rule feedback, use a tool for visual exploration, or connect to the broken workstation.

\subsection{Interface for navigation}

\begin{table*}[t]
    \centering
    \begin{tabular}{@{}cc@{}}
        \raisebox{-\height}{
        \begin{subfigure}[c]{0.79\textwidth}
        \includegraphics[width=\textwidth]{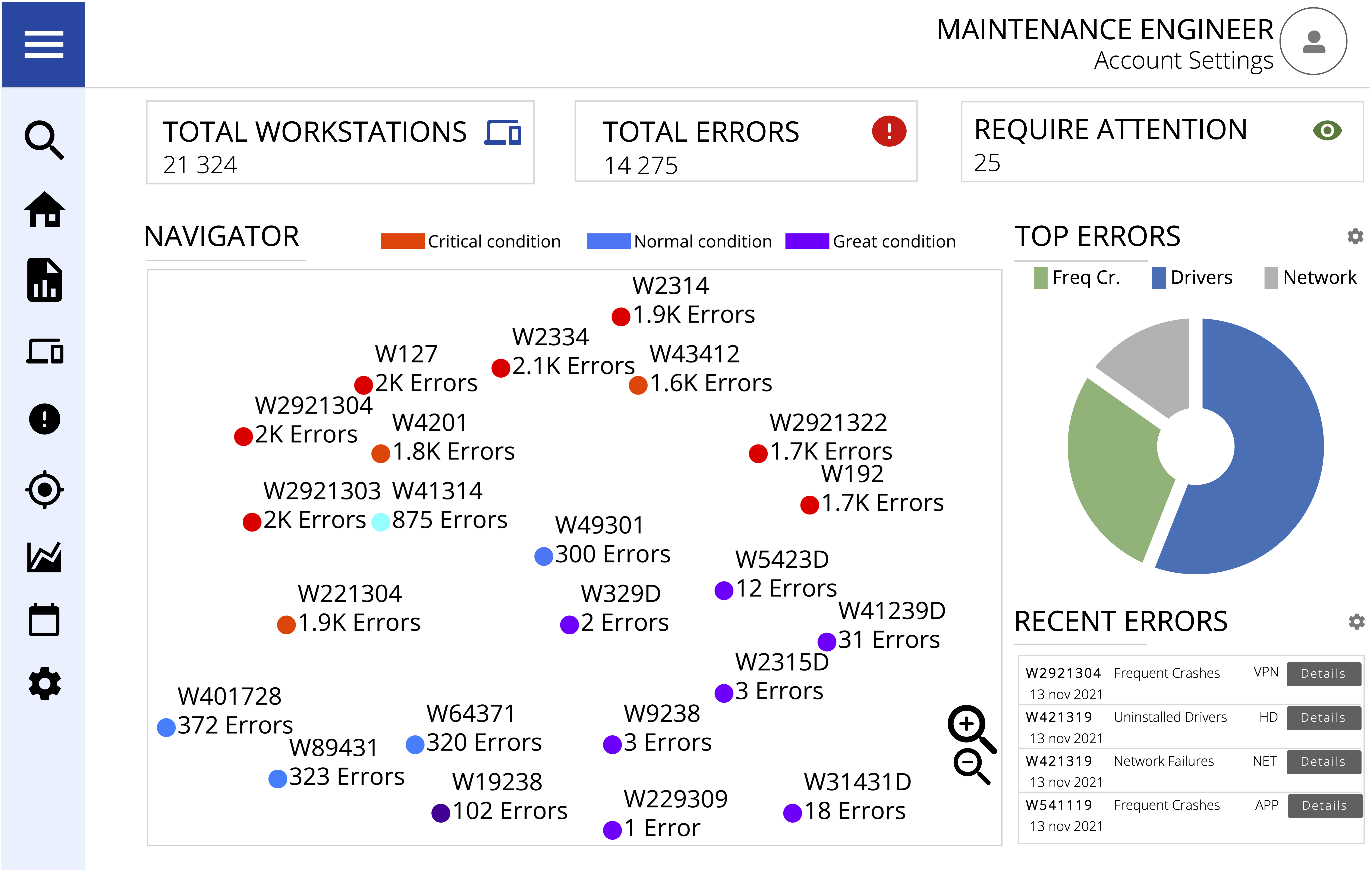}
        \caption{The navigation interface (mockup version). Users zoom in the visualized space of the alerts and explore areas with a high density of problematic workstations.}
        \end{subfigure}
        } & 
        \begin{tabular}[t]{@{}cc@{}}
            \raisebox{-\height}{
                \begin{subfigure}[c]{0.2\textwidth}
                \includegraphics[width=1\textwidth]{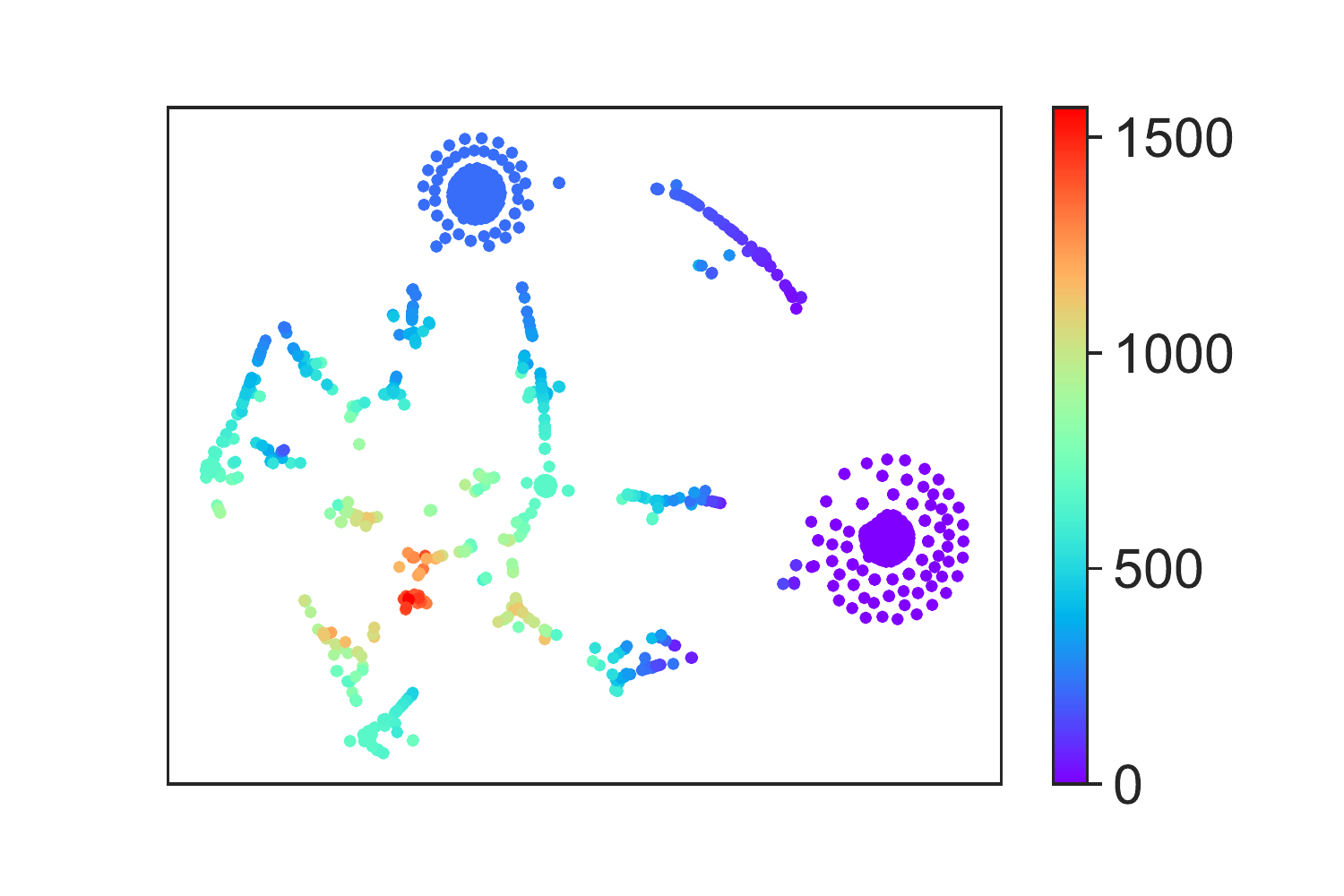}
                \caption{Company \#1.}
                \label{figure:navigation_interface_b}
                \end{subfigure}
            } \\[0.01cm]
            \raisebox{-\height}{
                \begin{subfigure}[c]{0.2\textwidth}
                \includegraphics[width=1\textwidth]{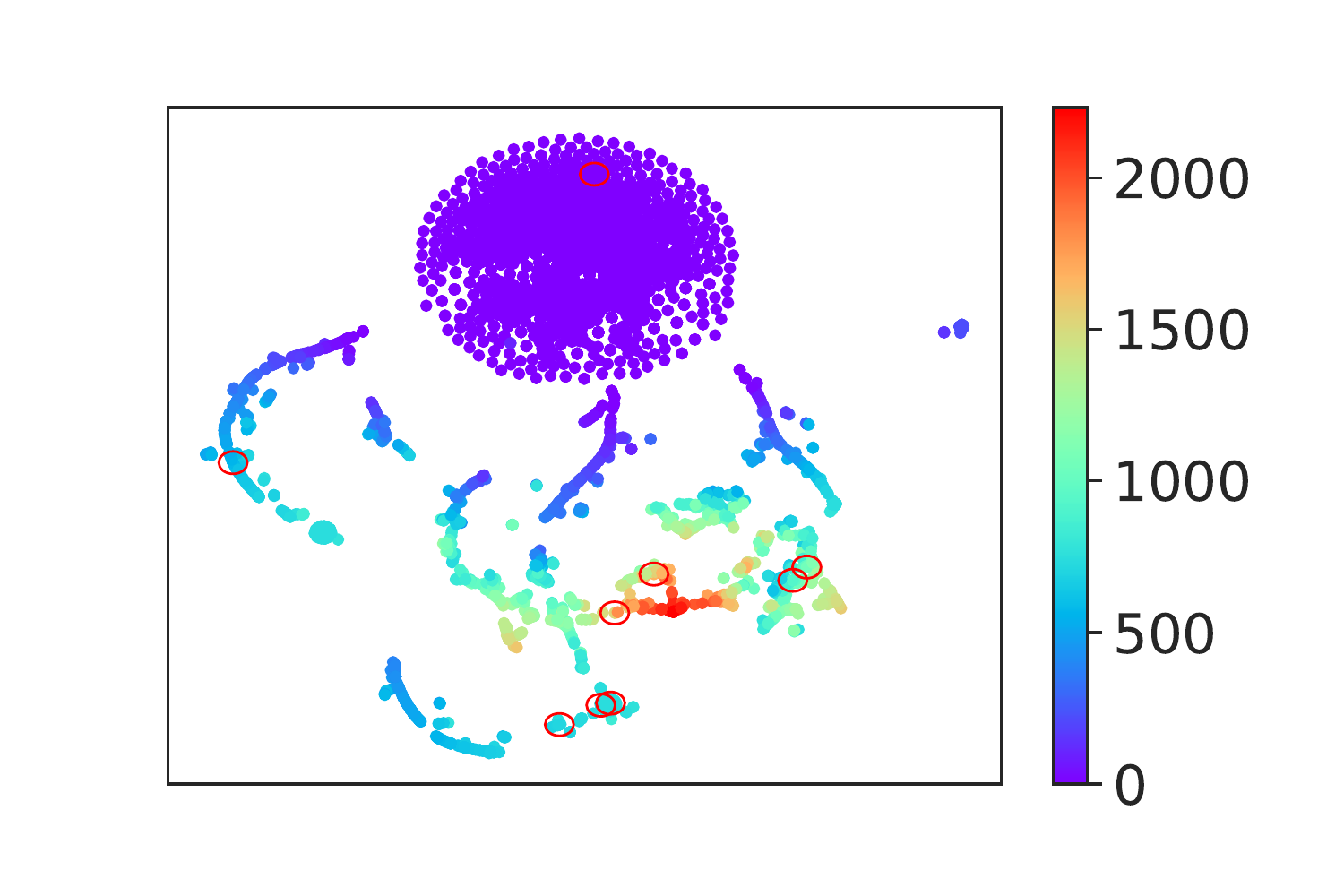}
                \caption{Company \#2.}
                \label{figure:navigation_interface_c}
                \end{subfigure}
            } \\[0.01cm]
            \raisebox{-\height}{
                \begin{subfigure}[c]{0.2\textwidth}
                \includegraphics[width=1\textwidth]{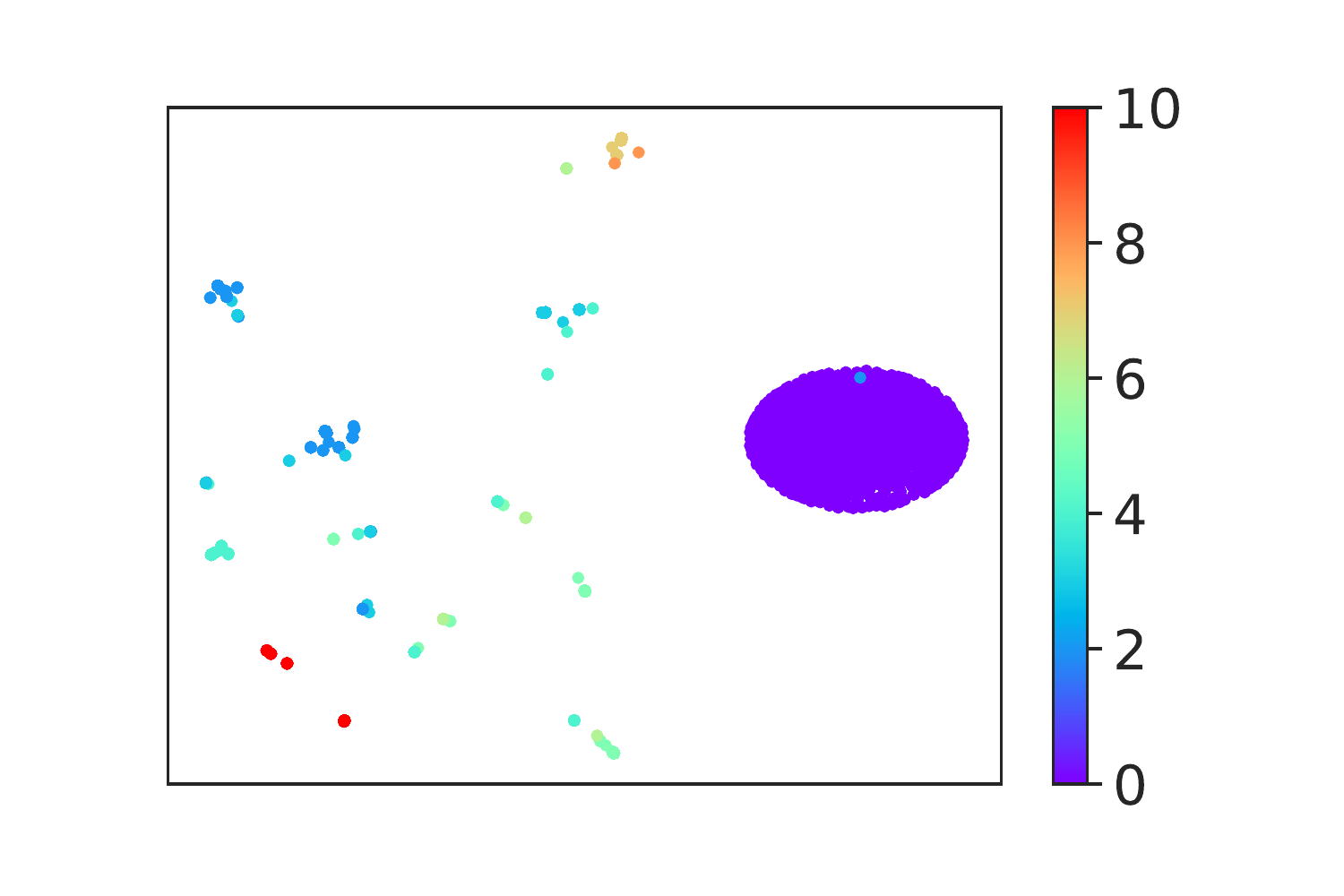}
                \caption{Company \#3.}
                \label{figure:navigation_interface_d}
                \end{subfigure}
            } & 
        \end{tabular}
    \end{tabular}
    \captionof{figure}{The interface for browsing the alerts in the visualized space of workstations. Image (a) shows the interface where workstations are the points. Each workstation is marked with colors that depend on how many alerts were collected for the workstation (in all figures). Navigating this space allows domain experts to identify similar workstations and fix them. The interface also shows the most frequent problems in the set of workstations as a pie chart on the right. Furthermore, in the right bottom corner, it shows the latest errors with a possibility to further investigate the issues. The visualization in (a) is a zoomed-in version of all workstations, with full plots of three companies shown in (b)-(c). The plots have been made with t-SNE \cite{van2008visualizing} which places similar data points close to each other and reveals clusters. Based on the coloring, clusters with problematic workstations can be easily identified as the red clusters in the images.}
    \label{figure:navigation_interface}
\end{table*}
\setcounter{table}{0}
    
An additional feature of the developed system is the possibility to visually navigate over the space of the workstations. The navigation process helps systematically identify problematic workstations and fix them with long-term measures. Fig.~\ref{figure:navigation_interface} shows the mockup version of the interface for domain experts and t-distributed stochastic neighbor embedding (t-SNE) for three different companies. For the second company, we also show workstations identified as chronically problematic by domain experts with red circles. Having this information, domain experts can zoom in on the visualizations in the areas close to systematically problematic workstations and explore similar workstations.

\subsection{Monitoring}
The deployed model has to be monitored. In our case, the predictions are provided to the domain experts daily, so one applicable metric is the number of valuable tickets (actionable or marked by domain experts as useful) per time interval. When this metric starts to fluctuate, it is a strong signal of a problem with the model, and either the model should be re-trained, the parameters should be adjusted, or the experts should be notified.

\section{Experiments}
We performed offline and online (with real customers) validation of the deployed service. For offline validation, we measured the model's performance on simulated and historical data (collected from customers' workstations). During online validation, we surveyed domain experts about real-life problems predicted by the proposed model.

\subsection{Offline experiments}
\label{section:offline_experiments}

\subsubsection{Data simulator}
\label{section:data_simulator}

We developed a synthetic data simulator based on survival analysis to generate an artificial dataset. The simulator facilitates reproducibility and provides a controllable environment for PdM experiments.

In the simulator, we assume that for each piece of equipment, a survival function $s(\lambda_i, t)$ exists. The survival function is a function that shows the probability of flawless performance (\say{survival}) of the $i$-th workstation up to time $t$. We use $s(\lambda_i, t) = exp(-\lambda_i t)$, where $\lambda_i$ is a constant hazard function specific for each workstation. We also introduce a parameter $\rho_i$ that shows the delay between experiencing problems and recovery. We assume that there are $N$ workstations, $D$ features describing each of them, and $T$ timestamps (in our experiments, $N=1000$, $D=12$, and $T=100$). We model both categorical and continuous features.

\textbf{Continuous features.} We assume that each continuous feature has two modes of functioning: normal and abnormal. We define features $x_{i}^{d, t}$ for the $i$-th workstation, $d$-th feature, and $t$-th timepoint as
\begin{equation}
\begin{aligned}
    x_{i}^{d, t} &=   \gamma(3, \lambda_i, t) \mathcal{N}_{\text{+}} + (1 - \gamma (3, \lambda_i, t)) \mathcal{N}_{\text{-}},\\
    \gamma(a, \lambda, t) &= \frac{a^{s(\lambda, t - t_r)} - 1}{a - 1},
    \mathcal{N}_{\text{+}} = \mathcal{N}(\mu_{\text{+}}, \sigma_{\text{+}}), \mathcal{N}_{\text{-}} = \mathcal{N}(\mu_{\text{-}}, \sigma_{\text{-}}), 
\end{aligned}
\end{equation}
where $t_r$ is the largest recovery time $\leq t$, and $\gamma$ is a balancing coefficient between normal and abnormal behaviour. Parameters $\mu$ and $\sigma$ are sampled from some distributions (these distributions are parameters of the simulator). This approach is motivated by exploring the behavior of the features in the real system: informative features tend to change their absolute values towards breaks.

\textbf{Categorical features.} Categorical features also contained two modes: normal and abnormal. Close to the point of malfunction, categorical variables switch from normal to abnormal mode. The speed of the switch is regulated by $\lambda_i$ and $\rho_i$ parameters and a stochastic component.

\begin{figure}[!htbp]
    \includegraphics[width=\columnwidth]{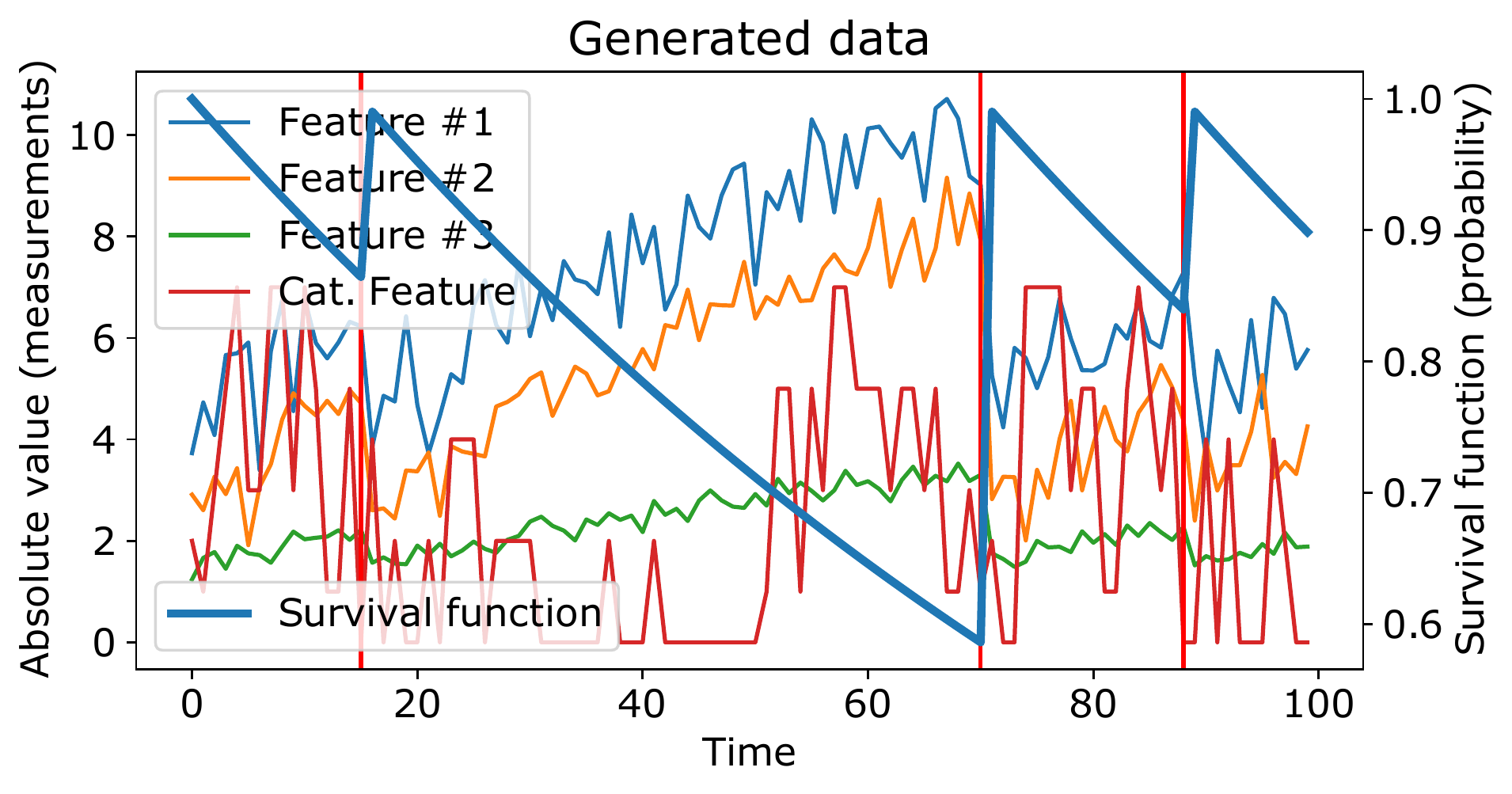}
    \caption{Generated data for one workstation. It is evident that three generated features increase toward the point of repair (marked with red vertical lines) and, then, restore their values.} 
    \label{figure:generated_data}
\end{figure}

In total, we modeled the distribution with eight categorical features and four continuous features. The generated data for one workstation is visualized in Fig.~\ref{figure:generated_data}.

We performed an experimental comparison of the methods described earlier in this paper. We first compared the ML methods and observed the following results: logistic regression ($f_1\text{-score}=0.2581 \pm 0.0121$), extremely randomised trees ($f_1\text{-score}=0.2796 \pm 0.0234$), and gradient boosting ($f_1\text{-score}=0.3425 \pm 0.0229$). By regulating signal to noise ratio via the simulator's parameters the results can be changed. We observe that the ranking of the models evaluated on synthetic data is similar to evaluated on historical data in Sec.~\ref{section:collected_data}, which shows the practical utility of the simulator.

We also modeled artificial domain experts via decision trees with low depth. Each human was parameterized with experience (the fraction of seen data from the training dataset) and depth (the complexity of decision rules produced by the expert), and included them as it is described in Sec.~\ref{section:dre}. We used $\text{depth} \sim \mathcal{U}(2, 8)$ and $\text{experience} \sim \mathcal{N}(0.75, 0.05)$. Decision rule elicitation improved $f_1\text{-score}$ to $0.5654 \pm 0.0202$. 

Our experiments indicated that with the growth in the number of feedback rules or the growth in the number of modeled domain experts, the predictive performance improves (see Appendix).

\subsubsection{Historical data}
\label{section:collected_data}
\begin{figure}
    \includegraphics[width=\columnwidth]{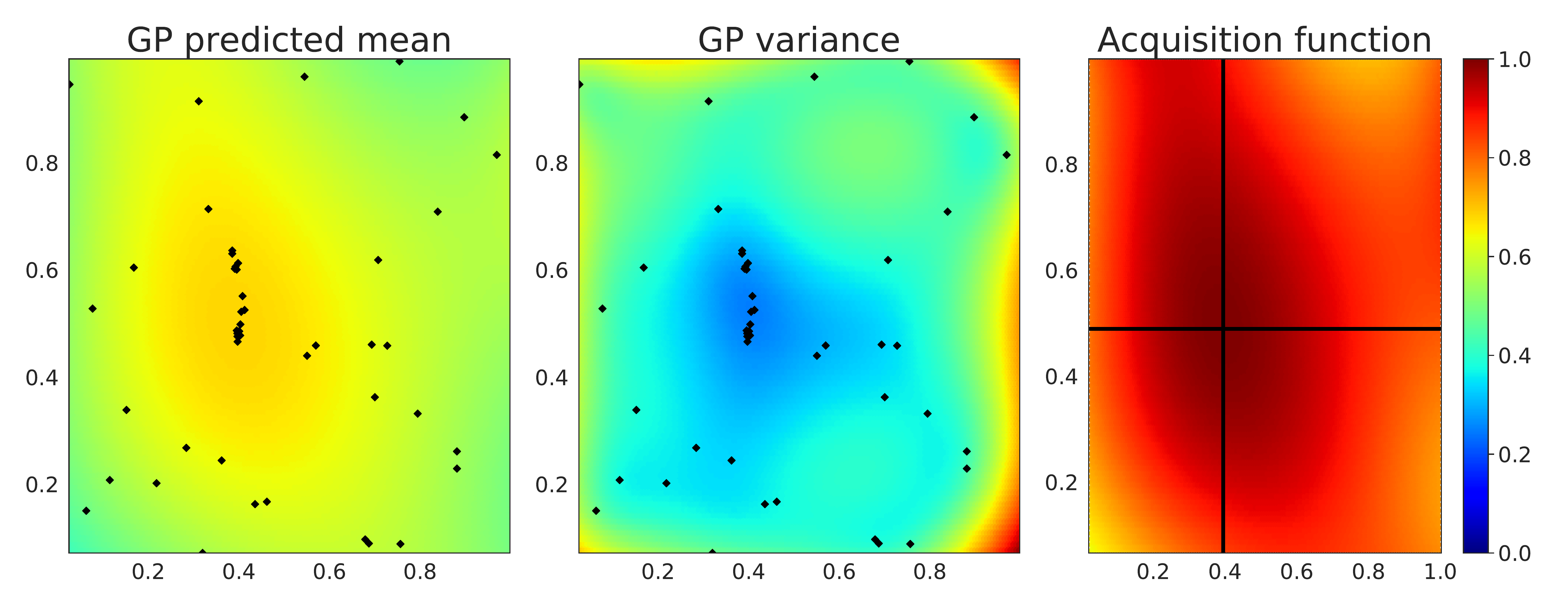}
    \caption{Bayesian optimization of the information and effect periods. Here, the axes are normalized information and effect periods, and the target function is the $f_1$-score. The colors show the GP mean on the left, the variance in the middle, and the acquisition function on the right. The acquisition function plot shows the next selected point during BO.} 
    \vspace{-1.5em}
    \label{figure:bo_iteration}
\end{figure}

\begin{table*}[htbp!]
    \centering
  \caption{Offline Experiments (10\% threshold).}
  \label{tab:offline_experiments}
  \begin{tabular}{lcccccl}
    \toprule
    Algorithm & Inf. period & Eff. period & $\text{f}_1$-score & precision & recall\\
    \midrule
    Logistic regression & 48h & 48h & $0.53 \pm 0.01$ & \bm{$0.87 \pm 0.02$} & $0.38 \pm 0.01$\\
    ExtraTrees classifier & 48h & 48h & $0.65 \pm 0.03$ & $0.58 \pm 0.03$ & \bm{$0.73 \pm 0.02$}\\
    Gradient boosting & 48h & 48h & $0.75 \pm 0.01$ & $0.84 \pm 0.02$ & $0.68 \pm 0.02$\\
    DRE (3) & 48h & 48h & $0.75 \pm 0.01$ & $0.86 \pm 0.02$ & $0.67 \pm 0.02$\\
    DRE (5) & 48h & 48h & $0.76 \pm 0.01$ & $0.84 \pm 0.02$ & $0.70 \pm 0.02$\\
    \textbf{DRE (20)} & \textbf{48h} & \textbf{48h} & \bm{$0.78 \pm 0.01$} & $0.85 \pm 0.02$ & \bm{$0.73 \pm 0.02$}\\
    \midrule
    Logistic regression & 31.9h & 119.9h & $0.51 \pm 0.02$ & $0.88 \pm 0.02$ & $0.35 \pm 0.01$\\
    ExtraTrees classifier & 31.9h & 119.9h & $0.76 \pm 0.01$ & $0.72 \pm 0.02$ & \bm{$0.82 \pm 0.01$}\\
    Gradient boosting & 31.9h & 119.9h & $0.77 \pm 0.01$ & $0.87 \pm 0.02$ & $0.69 \pm 0.02$\\
    DRE (3) & 31.9h & 119.9h & $0.77 \pm 0.01$ & \bm{$0.89 \pm 0.02$} & $0.67 \pm 0.02$\\
    DRE (5) & 31.9h & 119.9h & $0.78 \pm 0.01$ & $0.87 \pm 0.02$ & $0.71 \pm 0.02$\\
    DRE (15) & 31.9h & 119.9h & $0.80 \pm 0.01$ & $0.87 \pm 0.02$ & $0.74 \pm 0.02$\\
    \textbf{DRE (20)} & \textbf{31.9h} & \textbf{119.9h} & \bm{$0.81 \pm 0.01$} & $0.84 \pm 0.02$ & $0.77 \pm 0.02$\\
    \bottomrule
\end{tabular}
\end{table*}

We collected data from 1075 employee workstations at an undisclosed company for 21 days in total. The data were collected bi-hourly and contained the workstation measurements required for ML modeling and information about alerts, breaks, and user complaints.

We applied Bayesian optimization (as described in Sec.~\ref{section:bayesian_optimization}) with the UCB acquisition function, the RBF kernel for the Gaussian process, $\kappa=4$, and the decay coefficient $\kappa_{decay} = 0.99$. We used 30 initialization points and ran the optimization process for 20 iterations (however, we observed the convergence earlier). As the target function, we used the mean of $\text{f}_1$-score estimated by 10 evaluations with different train/test splits. The gradient boosting approach showed the best results across all the models; we found it to be consistently better than other models across the different values of $T_{\text{inf}}$, $T_{\text{eff}}$, and the threshold values; thus, we used it to estimate the quality of the proposed values of $T_{\text{inf}}$ and $T_{\text{eff}}$. As a result of BO, we obtained optimal values for the information and effect periods: $T_{\text{inf}} = 31.9h$ and $T_{\text{eff}} = 119.9h$. One of the Bayesian optimization iterations is visualized in Fig.~\ref{figure:bo_iteration}.

Next, we evaluated the performance of the machine learning approaches described in Sec.~\ref{section:ml_for_workstations_pdm}. We measured the $\text{f}_1\text{-score}$, the precision and recall of the algorithms and relied on the $\text{f}_1\text{-score}$ in the model selection process. We split 30\% of the dataset as a hold-out test set and performed test evaluation ten times with different random test sets and random seeds to obtain the confidence intervals. We also checked the results for different thresholds by the total number of alerts: 1\%, 5\%, and 10\%. We found that the models' $\text{f}_1\text{-score}$ was more than 0.7 for the 5\% and 10\% thresholds, with a drop to 0.52 for 1\%.

The experimental results are presented in Table~\ref{tab:offline_experiments}. DRE(n) is the proposed decision rule elicitation method applied to the gradient boosting model over decision trees with $n$ feedback rules. The experiment shows that decision rule elicitation improves the performance ($f_1$-score), and the improvement grows with the number of the feedback rules. The results are consistent between randomly selected information and effect periods and optimal values obtained via Bayesian optimization. The results (not absolute values but the ordering of the models) are also consistent with the simulator experiments of Sec.~\ref{section:data_simulator}.

\subsection{Online experiments}
\begin{figure}[!htbp]
    \includegraphics[width=\columnwidth]{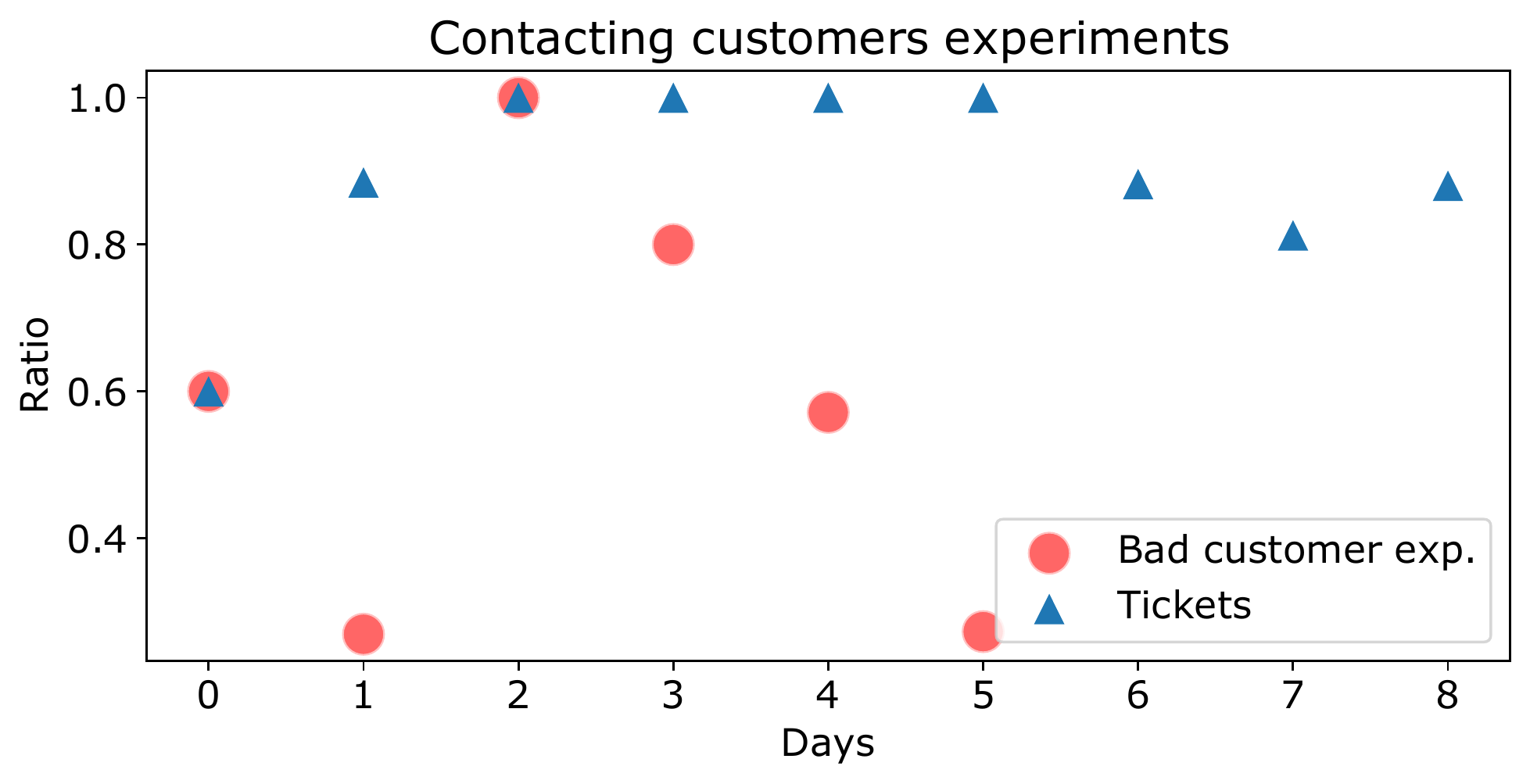}
    \caption{Online experiments. For randomly selected nine days over three weeks, we performed: \emph{(i)} analysis on how many predictions resulted in a ticket creation (domain experts were able to identify a problem, \mytriangle{blue_online}) and \emph{(ii)} how many predictions matched negative user experience (\protect\tikz\protect\draw[red,fill={rgb,255:red,255; green,101; blue,104}] (0,0) circle (0.5ex);). We can see that the majority of explored predictions resulted in a ticket; also, significant proportions of those were connected to the negative customer experience. The experiments were conducted for eight days (a longer period could cause a major overload on domain experts and affect customers). The total number of investigated tickets varied from three to 26 tickets per day.} 
    \label{figure:online_experiments}
\end{figure}
To validate the system in a production environment, we trained the model on one-month data. Then, our algorithm generated the predictions for domain experts for one month. Our analysis demonstrated that the predictive performance of the algorithm was consistent with offline testing --- the predictive performance ranking of the selected algorithms remained the same. After this, we selected several days and checked the predictions with customers to verify that the problems affected customers' experience during $T_\text{eff}$. We evaluated these predictions with domain experts and customers for nine days. The number of checked tickets and days per week was not controlled.

The results of the online validation are presented in Fig.~\ref{figure:online_experiments}. The validation shows the practical usefulness of the proposed approach: the average number of user requests per day measured for the last 2.5 years for this company was approximately six, which, compared to the results (with more than three tickets per day), promises a long-term reduction in the number of tickets of up to 50\%. Currently, the proposed system has been deployed for a year without technical issues and used by domain experts, which shows the validity of system design and modeling choices.

The validation took approximately four human hours per week, which included contacting the customers. We argue that the approach scales well with the growth in the number of workstations because the number of domain experts needed to support operations scales linearly with the number of devices (we assume that network maintenance is a separate task).

\section{Conclusion}
\label{section:conclusion}
This work presents a scalable machine learning approach to predictive workstation maintenance. The key feature of our approach is close interaction between human experts and machine learning algorithms. The synergy between expert knowledge and artificial intelligence leads to a scalable and robust method for modeling future failures in computer equipment. All the steps were designed to effortlessly generalize the approach to other domains or more equipment, leading to accessible horizontal and vertical scaling. We demonstrated the implementation of our approach to a production case and reported experimental results. Even though we considered only workstation PdM problems, our system can be applied to other PdM tasks without drastic changes.

Our HITL approach was successful both with historical data and in practice. The combination of ML models and decision rule feedback improved the results in the experimental setup and allowed experts to influence the model results. During the online validation, we detected and fixed the problems that affected the users; the observed performance is sufficient to reduce the total number of user complaints by up to 50 percent in the long term. The notification system has been working for one year, which proves the effectiveness of chosen design and implementation.

Our system is a significant advance towards automated, reliable problem management, which could impact a large variety of business sectors.

\begin{acks}
This work was supported by the Academy of Finland (Flagship programme: Finnish Center for Artificial Intelligence FCAI) and UKRI Turing AI World-Leading Researcher Fellowship, EP/W002973/1. We also acknowledge the computational resources provided by the Aalto Science-IT Project from Computer Science IT.
\end{acks}

\bibliographystyle{ACM-Reference-Format}
\bibliography{main}

\appendix

\newpage
\section{Reproducibility}
\subsection{Synthetic Data}
\label{appendix:synthetic_data}

The simulator is suitable for performing controllable experiments. For example, in some systems, the recovery can be faster than in others, and by tunning parameter $\rho$, developers can produce data that resembles a specific system. The stochasticity of parameters $\lambda$ and $\rho$ models how robust to problems different workstations are.

The parameters of the generator are related to the real system's properties in Table~\ref{table:simulator_params}. There we denote as $D_{+}$ and $D_{-}$ categorical variable distributions for normal and abnormal functioning. We manually tuned the simulator's parameters presented above to approximately reflect the characteristics of the workstations. The data-driven search of simulator parameters can be performed by, for example, approximate Bayesian computation \cite{csillery2010approximate}.

\vspace{1em}
\begin{table}[!htbp]
      \caption{Simulator parameters and real system counterparts.}
      \begin{tabular}{lll}
        \toprule
        Variable & System & Value\\
        \midrule
        $s(\lambda_i, t)$ & degradation speed & $exp(-\lambda_i t)$ \\[0.2cm]
        $x_i^{t, d}$ & quantitative characteristics &\shortstack{ $\mathcal{N}_{\text{+}}, \mathcal{N}_{\text{-}}$ \\ $\mathrm{D}_{\text{+}}, \mathrm{D}_{\text{-}}$}\\[0.1cm]
        $\rho_i$ & delay/complexity of repair & $\Gamma(1, 0.1)$\\
        \bottomrule
      \end{tabular}
    \label{table:simulator_params}
\end{table}
\vspace{1em}

\emph{Parameters.} We use the following generation process for continuous features:
\begin{equation}
\begin{aligned}
    \lambda_i &\sim \Gamma(1, 0.005), \rho_i \sim \Gamma(1, 0.1),  \\
    p(\text{rec.}) &= 1 - s(\lambda_i, t - t_r) - \rho_i,  \\
    \mu^{d}_\text{+} &\sim \Gamma(2, 1), \sigma^{d}_{\text{+}}, \sim \Gamma(1, 1),  \\
    \mu^{d}_{\text{-}} &\sim \mu^{d}_{\text{+}} + \Gamma(4, 2), \sigma^{d}_\text{-} = 1.5 * \sigma^{d}_{\text{+}}, \\
\end{aligned}
\end{equation}
where $p(rec.)$ models probability of the system's recovery. For categorical features, we sample a variable $z \sim U[0, 1]$, if $z \leq 1 - s(\lambda_i, t) - \rho_i$ we consider a workstation broken, and if $z \leq 1 - s(\lambda_i, t)$ the switch to abnormal functioning mode occurs.

We acknowledge that for better reproducibility of real workstations, it might be useful to add uninformative features; however, we found using only informative features sufficient for our case because domain experts selected the highly relevant features in real data. Or adjusting the distributions of the parameters to better mimic the data.

Here, we show the predictive performance of DRE methods on a synthetic dataset. We observe an improvement in predictive performance when the number of experts grows. As the base model, we have used gradient boosting over decision trees, and 95-\% confidence intervals were extracted by running experiments on different random seeds and train/test split for 20 iterations.

\begin{figure}[!htbp]
    \includegraphics[width=\columnwidth]{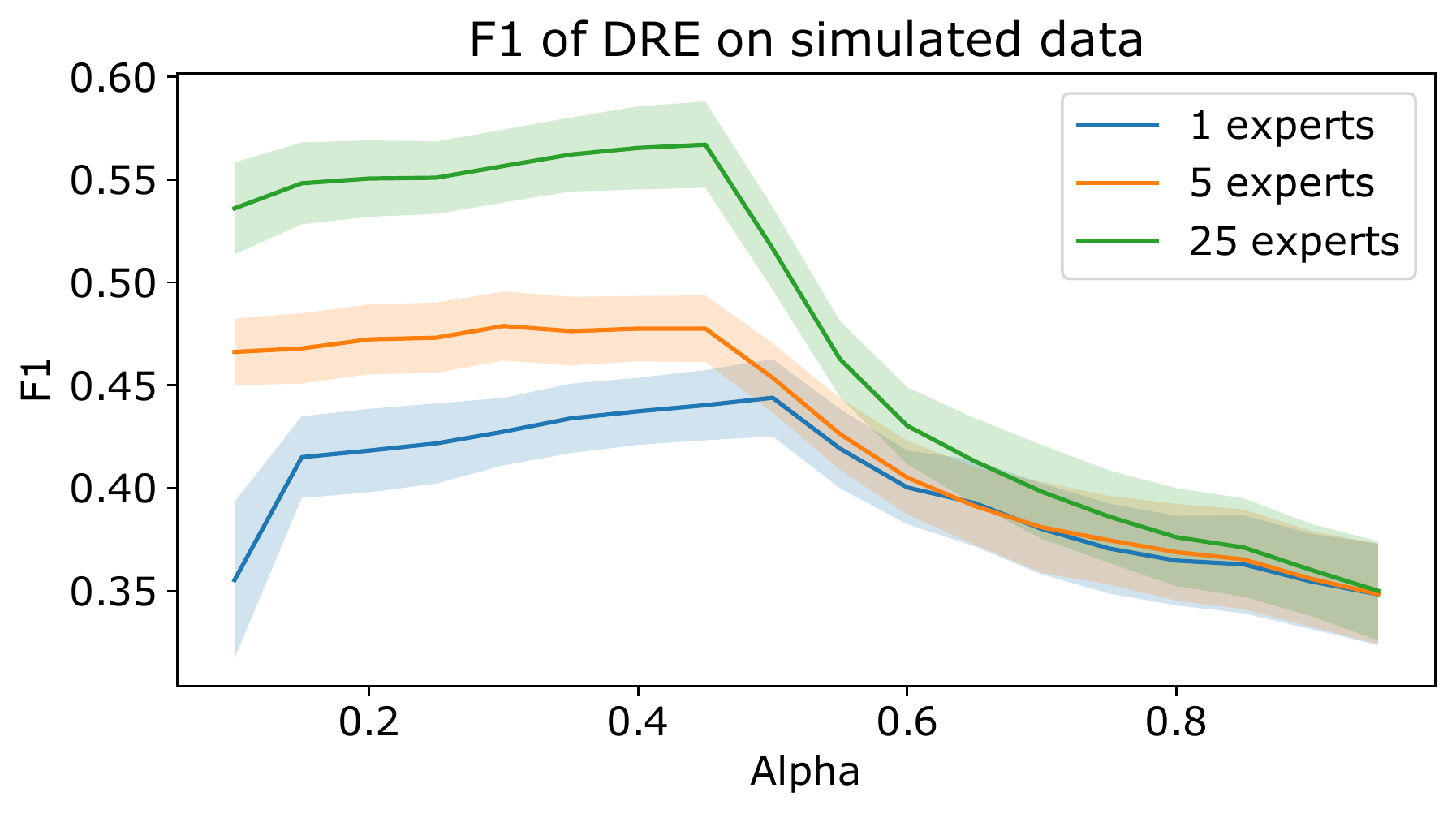}
    \caption{The graph shows the predictive performance of the DRE classifiers. We observe that when the number of domain experts grows, the predictive performance is improved. Also, an optimal value of $\alpha$ inside the interval shows that the combination of feedback rules and a data-driven model shows the best results.} 
    \label{figure:f1_dre_humans}
\end{figure}

\subsection{Historical Data Experiments}
To collect historical data, one needs to install agents for each workstation and continuously monitor a variety of measurements. It can be done, e.g., by reading system monitors, system logs or using OS-specific utilities (e.g., top and uptime in Unix). The agents submit their measurements to a central server responsible for storing the data. More specifically, the measurements should collect data regarding CPU performance, hard drive work, information about all actions with drivers (e.g., installation, update, and removal), and information about system errors and warnings. The collected measurements can be extended to a wider variety of characteristics, e.g., network performance measurements or information about the system's temperature.

Hyperparameters of the method can be found by the Bayesian optimization, as described in Sec.~\ref{section:bayesian_optimization}. We do not provide specific hyperparameters in this work, as it can significantly change from vendor to vendor and other unobserved causes and thus does not contain a practical value. However, we saw that this approach is extendable to different domains (e.g., different vendors), as we tested the proposed approach in application to four companies and observed practically beneficial predictive performance results with all of them.

\end{document}